# Copebot: Underwater soft robot with copepod-like locomotion

Zhiguo He[a,b,1,∇], Yang Yang[a,∇], Pengcheng Jiao[a,b,1], Haipeng Wang[a], Guanzheng Lin[a], Thomas Pähtz[a]

a: Institute of Port, Coastal and Offshore Engineering, Ocean College, Zhejiang University, Zhoushan 316021, Zhejiang, China
b: Engineering Research Center of Oceanic Sensing Technology and Equipment, Zhejiang University, Ministry of Education, China

**Abstract**

It has been a great challenge to develop robots that are able to perform complex movement patterns with high speed and, simultaneously, high accuracy. Copepods are animals found in freshwater and saltwater habitats that can have extremely fast escape responses when a predator is sensed by performing explosive curved jumps. Here, we present a design and build prototypes of a combustion-driven underwater soft robot, the "copebot", that, like copepods, is able to accurately reach nearby predefined locations in space within a single curved jump. Because of an improved thrust force transmission unit, causing a large initial acceleration peak (850 Bodylength·$s^{-2}$), the copebot is 8 times faster than previous combustion-driven underwater soft robots, whilst able to perform a complete 360° rotation during the jump. Thrusts generated by the copebot are tested to quantitatively determine the actuation performance, and parametric studies are conducted to investigate the sensitivities of the input parameters to the kinematic performance of the copebot. We demonstrate the utility of our design by building a prototype that rapidly jumps out of the water, accurately lands on its feet on a small platform, wirelessly transmits data, and jumps back into the water. Our copebot design opens the way toward high-performance biomimetic robots for multifunctional applications.

**Keywords:** underwater soft robots; combustion actuation; fast and maneuverable locomotion; wireless communication

[1] Corresponding authors. Emails: pjiao@zju.edu.cn (P. Jiao) and hezhiguo@zju.edu.cn (Z He).
[∇] These authors contributed equally to this work.

## Introduction

Everyone who has seen the clumsy and slow movements of football-playing robots understands the challenge engineers are facing in developing robots that are able to mimic the complex movement patterns of living beings on land, let alone under water, with high speed and, simultaneously, high accuracy[1-9]. A particularly curious example of a complex movement pattern of an aquatic animal is the locomotion of copepods, which are creatures found in almost every freshwater and saltwater habitat[6]. Some copepods are able to use their tails and antennae (Figure 1(a)) so explosively that they are able to jump a distance of 500 times their body lengths within about 1 s[7]. Furthermore, while jumping, a copepod is able to evade predators by altering the direction of the jump as well as the orientation of its body in space using its other body parts (Figure 1(b)). The movability and maneuverability of copepods endow them with fascinating abilities to survive in harsh conditions. However, lack of underwater soft robots has been developed to perform similar/comparable motions under the current actuation methods. Taking advantage of the combustion actuation, the reported copebot can reproduce the transiently high-acceleration underwater motion and similar flexibility during high-speed motions.

To mimic these kinematic characteristics of copepods, here, we present a design of a biomimetic underwater robot - copebot. The copebot is designed based on the transient driving method[8]. Its jump driven by the combustion of a mixed gas of $O_2$ and $C_3H_8$ and consists of soft and flexible materials to keep its weight low and to allow an efficient thrust force transmission under water. The force transmission unit is much improved over those in previous combustion-driven underwater soft robots[10], allowing the copebot to move very explosively (850 Bodylength·$s^{-2}$ at the initial acceleration peak) and 8 times faster (roughly up to 4 body lengths per 500 ms). Furthermore, adjustable wings on the side of its torso, allow the copebot, like copepods, to change its jump direction and also its orientation in space via rotating by up to 360° while jumping. We have carried out experiments showing that the copebot is able to accurately reach a predefined nearby location even under challenging conditions. We have quantitatively tested the thrusts generated by the copebot and conduct the parametric studies to investigate the sensitivities of the input parameters to the kinematic performance. Moreover, locomotion experiments have been conducted to test the potential application of the copebot design for building signal-transmitting underwater robots, that is, robots that are able to send data through a multiphase environment (i.e., water-to-air). Numerical and theoretical studies have been carried out to analyze the performance of the copebot (Supporting Information). The significance of the reported biomimetic copebot can be summarized in movability, maneuverability and multifunctional applications:

1) **Movability**: The copebot can explosively move with the velocity of 850 body length per second squared at the initial acceleration peak, which is averagely 8 times faster than the existing underwater soft robots [10-11, 44-50]. The unprecedented high movability effectively mimics the "escape jump" of copepods for the first time in the literature.

2) **Maneuverability**: The copebot can be maneuvered to perform straight and steering jumps with satisfactory accuracy by mimicking copepods, which introduces a simple but promising method to control the extremely rapid locomotion of soft robots [10, 25-27].

3) **Multifunctional applications**: Benefiting from the well movability and maneuverability, the copebot can operate various complex tasks, such as leaping out of water to transmit signals collected underwater, rapid underwater attitude adjustment, and accurately reaching preset targets.

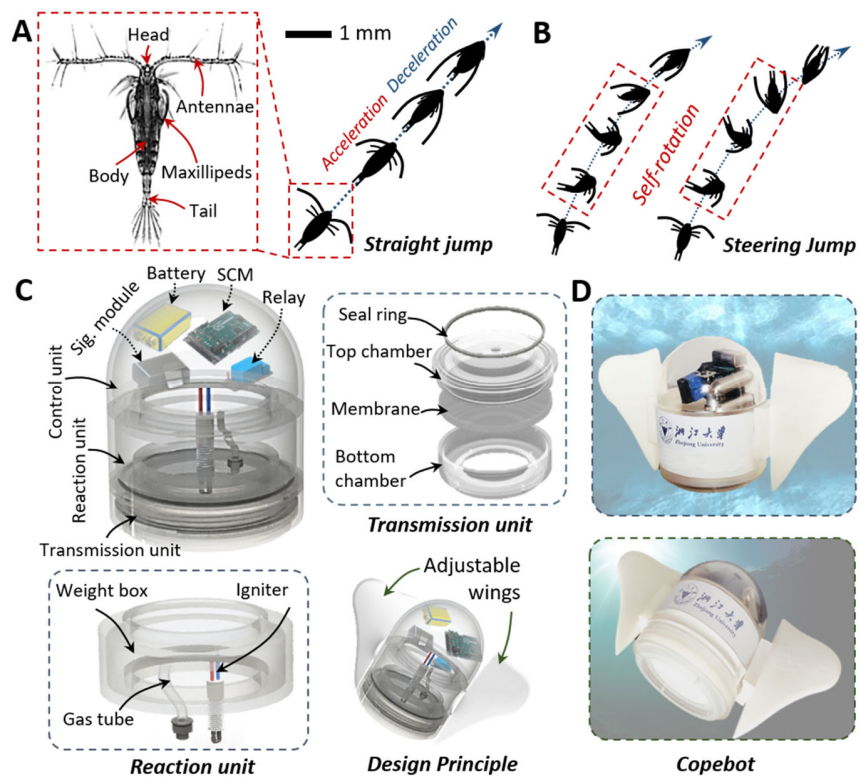

**FIG. 1.** Design principle and applications of the copebot. **(a)** Illustration of the copepod locomotion. The copepod mainly consists of a head, antennae, a body, maxillipeds, and a tail. **(b)** Illustration of the copepod locomotion. The copepod is able to perform steering jump with self-rotations: it can rotate and turn back to avoid predators and can rotate up to 360° during the jump. **(c)** Schematic of the copebot. **(d)** The photographs of a copebot prototype.

The design strategy in this study can be applied in developing soft robots for more challenging scenarios, including jumping in the weightless environment (e.g., space exploration), overcoming external disturbance (e.g., complex fluid field condition or rough terrain), and rapid obstacle avoidance (e.g., avoiding hunting or attack).

**Design principles and experimental setups**

*Design principles*

The design of the copebot consists of three major functional components: a control unit, reaction unit, and transmission unit, as illustrated in Figure 1(c). The function of the control unit, which contains a single-chip microcomputer (SCM), relay, battery, and signal module, is to activate the igniter and send and receive signals when the SCM posts instructions. The reaction unit, which consists of a weight box, igniter, and gas tube, is designed to provide the tools to generate combustions. The transmission unit is a well-sealed chamber where combustions take place. It contains a seal ring, top and bottom chambers, and an expandable membrane. The diameter, height, and weight of the copebot are 150 mm, 200 mm, and 2 kg, respectively. A series of different experiments were undertaken (Movies S1 and S2). The gas was premixed in the ratio of 4.8 (Oxygen:Propane) and was input into the copebot with speeds of 24 ml/min and 5 ml/min for 80 seconds. The recording devices were a high-speed optical capture system and a single lens reflex camera.

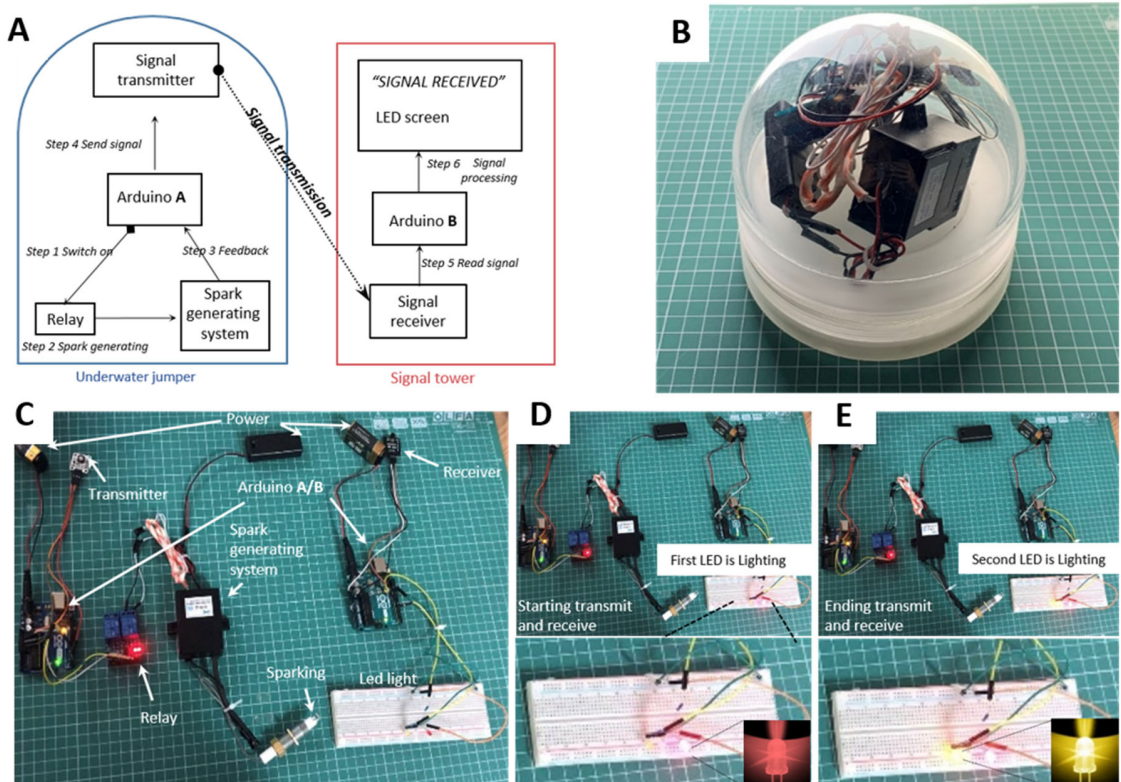

**FIG. 2.** The signal transmission system: **(a)** the schematic of the system design, **(b)** the assembled copebot, **(c)** the signal transmitting and receiving devices in the copebot, and **(d)** the transmitting and receiving process of the electrical signals.

*Experimental setups of the signal transmission experiment*

The design principle of the signal transmission system is shown in Figure 2. The control systems of the copebot are shown in the left part of Figure 2(a), including a signal chip microcomputer (SCM Arduino), rely, spark generating systems and signal transmitter. Ignition and transmission programs were preset into the SCM. These programs controlled the switch of the relay. The switch started the spark generating system activating the combustion within the specified time. After the combustion happened in the reaction unit, the feedback was sent back to SCM immediately to start up the signal transmitter unit. In the signal tower, SCM recognized the signal received by the signal receiver and displayed the signal on the LED screen. The reported transmitting process can capture the out-of-water moments and transmit the signal in time. The signal received from data source (e.g., a signal tower) can be processed by the SCM and reflect on the terminal (e.g., the LED screens). Figure 2(b) shows the assembly of the copebot. The experimental setup for signal transmission and the different working states of the system are shown in Figure 2(c). Three power supply units provide power to the signal transmitter part, the ignition part and the signal receiver part of the system. The signal transmitter part includes the SCM and the transmitter terminal. A relay and a high volatge pack are applied to generate sparks. The signal receiver part contains an SCM and a breadboard to control the receiver and connect to LED lights. The spark was automatically ignited and triggered the end of the copebot in preset time, as shown in Figure 2(c). The signal transmitter started delivering electrical signals, and when the signal receiver captured the information, the first LED bulb was lighted, as shown in the Figure 2(d). The second LED bulb was lighted when the data processing was finished, as shown in the Figure 2(e).

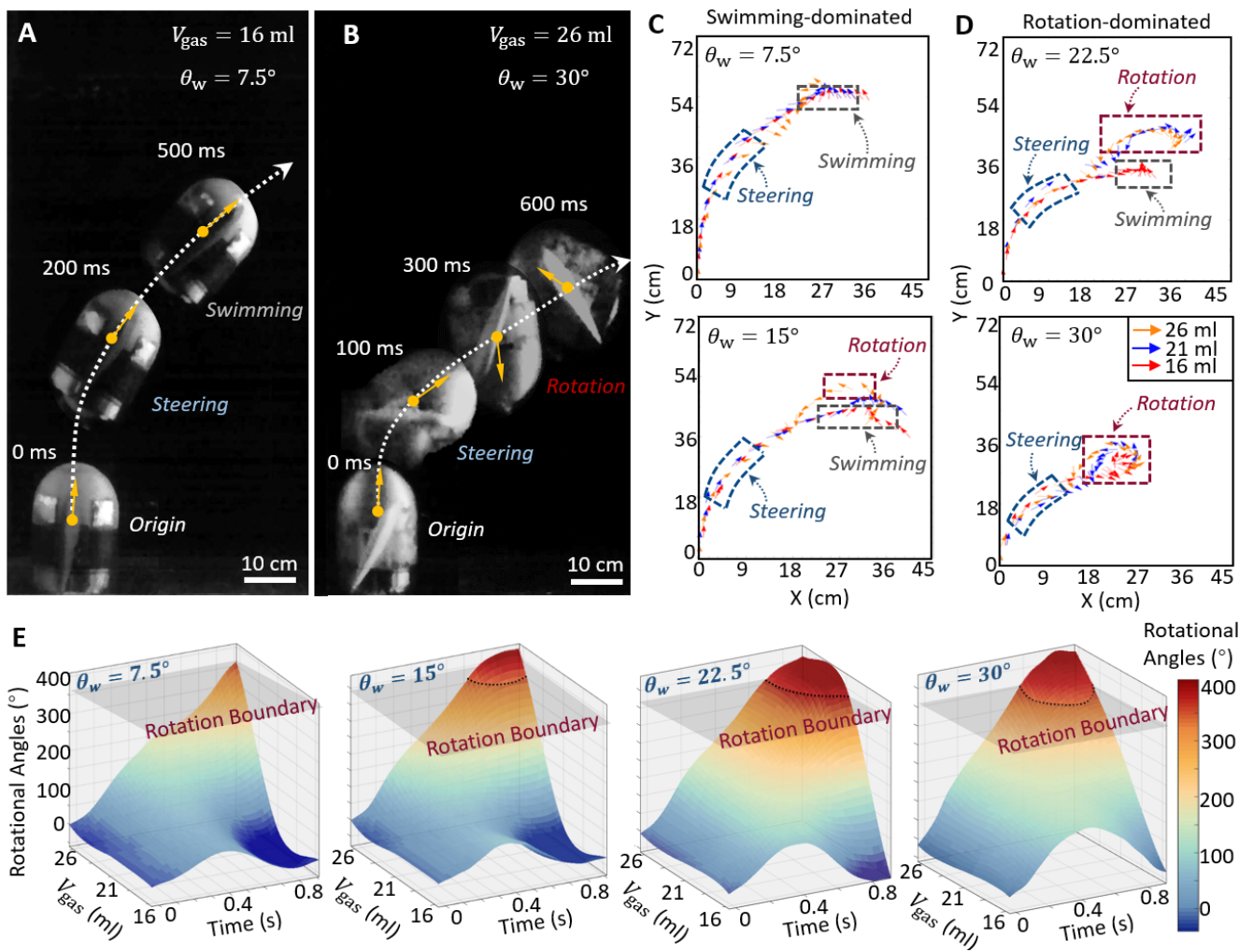

**FIG. 3.** Maneuverability experiments of the copebot by the adjustable wings. **(a)** Swimming-dominated motions obtained for a gas amount of 16 ml and wing angle of 7.5°. **(b)** Rotation-dominated motions for a gas amount of 26 ml and wing angle of 30°. **(c) & (d)** Motion traces of a copebot prototype captured from maneuverability experiments for the **(c)** swimming-dominated motions and (d) rotation-dominated motions. The locations of the arrows indicate the transient robot locations captured by a high-speed camera and processed using the binarization method. **(e)** Relations between the gas amount, time, and body rotations for different wing angles. The rotation boundary is the plane referring to a rotation of 360°, which has been defined to distinguish the swimming-dominated and rotation-dominated motions.

**Experimental results**

*Underwater maneuverability experiments.*

The underwater maneuverability experiments are conducted in a laboratory-grade testing flume with the width-length-height of $1.2 \times 1.2 \times 2.4$ m. Figure 1(d) shows that a pair of wings can be attached to a copebot's torso. Since the wings can be turned to different inclined positions, the copebot is able to jump straightly or with steering and rotation motions in a relatively short time period. To explore the controllability of the copebot, we have experimentally investigated the steering motions caused by different gas amounts and pre-inclined wing angles. Figures 3(a) and 3(b) show the motion images of a copebot prototype captured from the experiments for two exemplary cases corresponding to a swimming-dominated motion (Figure 3(a)) and rotation-dominated motion (Figure 3(b)). Note that the rotation motion in Figure 3(b) dissipates a lot of energy, causing the robot to travel a smaller distance than in Figure 3(a) in spite of exhibiting a large initial thrust (also see Movies S1 and S2). Considering the different characteristics of the swimming-dominated and rotation-dominated motions, the copebot design can be used for applications requiring rapid movements and for those requiring rapid head-orientation adjustment. Figures 3(c) and 3(d) present the motion traces of a copebot prototype in the swimming-dominated ($\theta_W = 7.5°$ and $\theta_W = 15$) and rotation-dominated ($\theta_W = 22.5°$ and $\theta_W = 30°$) motions, respectively. The robot is tested for 12 cases (i.e., 4 wing angles by 3 gas amounts). The arrows represent the transient head-orientations of the robot, which were obtained by binary processing results from the grayscale images captured by a high-speed camera. It can be seen that increasing the gas amount increases rotation because of increased fluid drag torque on the wings. Note that, in all the cases, the robot finished the motions with head-up postures (not shown), which we achieved through setting the gravity center of the robot lower than the buoyant center. Figure 3(e) shows the relations between the gas amount, time, and robot rotations with respect to different wing angles. We defined $\omega = 360°$ as the rotation boundary to distinguish the swimming-dominated and rotation-dominated motions. In the case of $\theta_W = 7.5°$, the rotational angles were not able to reach the rotation boundary even for gas amounts of 26 ml. In contrast, when the wing angles exceeded $\theta_W = 15°$, the copebot performed rotation-dominated motions for sufficiently large gas amounts.

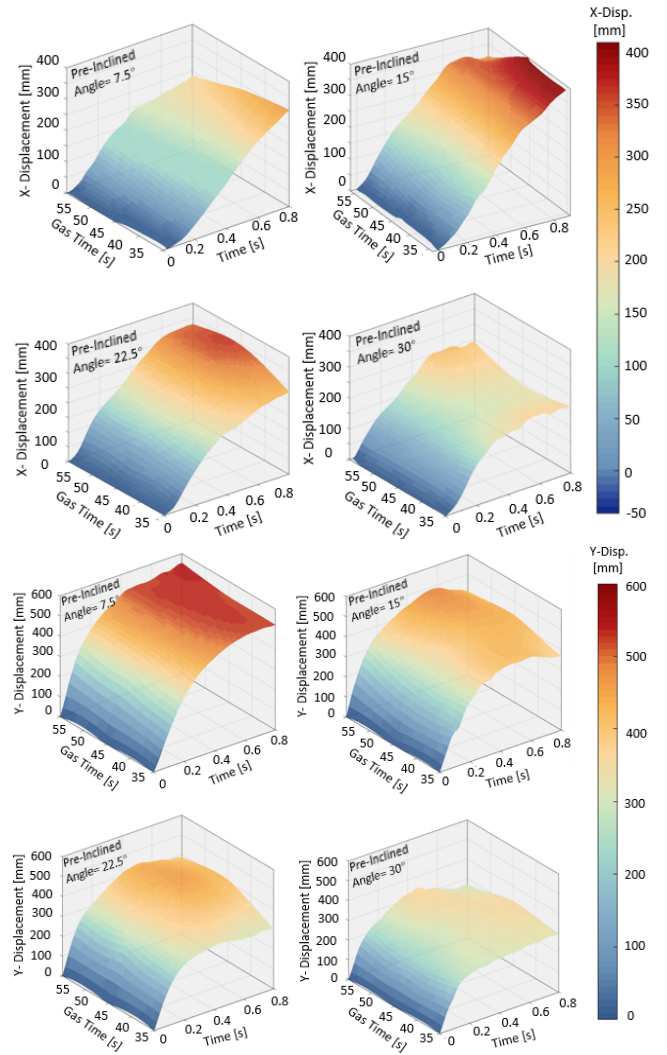

**FIG. 4.** Directional displacements of the copebot. It presents the displacements of X-direction and Y-direction respectively with different values of pre-inclined angles.

*Experimental analysis of the steering experiments for the copebot*

The steering experiments are conducted in the reported flume. By obtaining the optical images of the steering copebot, we got binary images by using threshold values and morphology processing. In steering experiments, the frame rate of the high-speed camera was set as 1000 FPs and the robot locations were obtained in every millisecond[11-27, 44]. In the traces of the copebot's motions, there are steering and rotation motions which are the results of the specific input parameter (i.e. gas amount and pre-inclined wing angles.). When the pre-inclined wing angles are relatively low (e.g. 7.5° - 35s, 45s and 55s) or the input gas amount are relatively low (e.g. 35s - 7.5°, 15° and 22.5°), the robots tend to operate steering motions which can fulfill propulsion requirements. Moreover, when the pre-inclined wing angles are relatively large (e.g. 30° - 35s, 45s and 55s) or the input gas amounts are relatively large (e.g. 55s - 15°, 22.5° and 30°), the robots tend to operate rotation motions which can meet fast posture adjustments requirements. The relations between robot traces and pre-inclined wing angles in different input gas time are shown in Figure 4. In steering motions, the horizontal displacement increases with the increment of pre-inclined wing angles due to the stable head orientation changing. However, in rotation motions, both horizontal and vertical displacements decrease with the increment of the pre-inclined wing angles due to the energy consumption of the robots tossing. Moreover, the vertical displacement increases with the increment of gas amounts. The same relations can also be seen in Figure 4. To quantitatively analyze the steering performance, the empirical formulas are developed as

$$i = t(\alpha_i + \beta_i + \gamma_i), \quad (1)$$

where

$$\begin{cases} \alpha_i = A_i \frac{g}{T_c} t^2 \\ \beta_i = \frac{\mu S}{m}(B_{i1} - B_{i2}\theta_w), \\ \gamma_i = \frac{C_i T_c^2}{S} V_{gas} \end{cases} \quad (2)$$

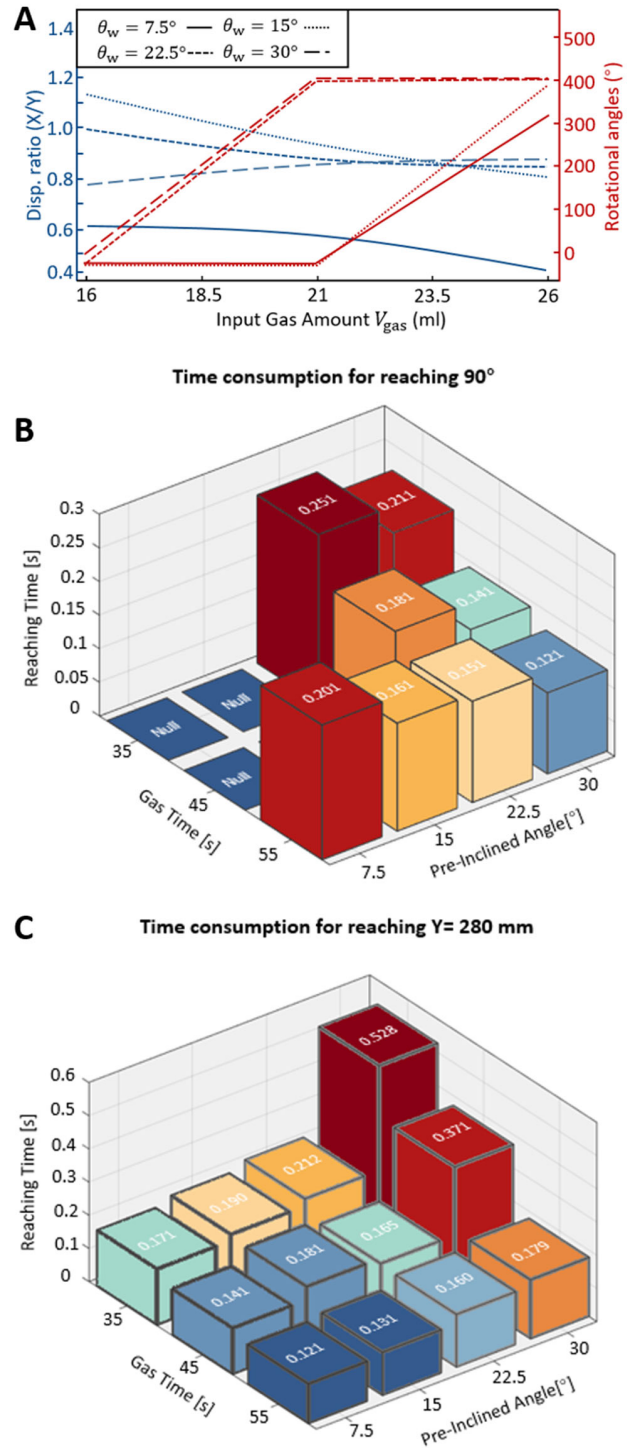

FIG. 5. **(a)** Relations between displacement directional ratio, rational angles and input gas amount. **(b)** Parametric studies for the time consumptions to reach 90-degree rotation. **(c)** Parametric studies for the time consumption to reach Y=280 mm.

TABLE 1. DETAILED VALUES OF EMPIRICAL PARAMETERS.

| Parameter | Value | Parameter | Value |
|---|---|---|---|
| $A_X$ | $-9.1 \times 10^{-11}$ | $B_{Y2}$ | $1.3 \times 10^5$ |
| $A_Y$ | $-1.91 \times 10^{-9}$ | $B_{Z1}$ | $1.3 \times 10^5$ |
| $A_Z$ | $-3.7 \times 10^{-10}$ | $B_{Z2}$ | 8475 |
| $B_{X1}$ | 659.7 | $C_X$ | 3.9 |
| $B_{X2}$ | 12747 | $C_Y$ | 12 |
| $B_{Y1}$ | 22033 | $C_Z$ | 3.1 |

where $g = 9.8$ m/s$^2$, $T_c$, $\mu$, $S$, $t$, $V_{gas}$, $\theta_w$ are the gravity, the time period of the combustion, the dynamic viscosity of the water, the effective acting area of the combustion, the mass of the copebot, time, the gas amount and the pre-inclined wing angle, respectively [28-43]. Note that $i$ represents X, Y, and $\omega$, which are the displacement in x-direction, the displacement in y-direction, the rotational angle, respectively. The detailed values of empirical parameters are shown in Table 1. Eqs. (1) and (2) are generalizable empirical formulas for summarizing motion performances of similar systems.

*Parametric studies.*

To explain the physical principles during the steering motions, we summarized the relations between the directional displacement ratio, rotational angles, and input gas amount under different pre-inclined wing angles, as shown in Figure 5(a). The energy (combustion released) was consumed by the horizontal motion, vertical motion, and head-orientation changing. Regarding the directional displacement ratio (i.e., X:Y), it started declining from 0.6 when the pre-inclined wing angle was 7.5°. When the pre-inclined wing angle was increased to 15°, the directional displacement ratio was raised to the highest value, which indicates that 15° was the optimal angle for the horizontal motion. To further develop the maneuverability of the copebot, systematic studies were conducted in Figures 5(b) and 5(c). We studied the time consumption to reach the head orientations of 90° and the height of Y=280 mm. To reach the 90° target, the time consumption reduced with increments of the pre-inclined wing angles and the gas amount. To reach the Y=280 mm target, the vertical motion dominated cases (i.e., 7.5°-55 s) performed the most accurately.

*Thrust testing.*

To quantitatively obtain the actuation performances of the copebot, the thrust combustion generated should be experimentally tested. Then, the obtained values can be applied as initial conditions of the further numerical and theoretical studies (see Supporting Information).

TABLE 2. GEOMETRIC PROPERTIES AND EXPERIMENTAL CONDITIONS OF THE COMBUSTION CALIBRATION.

| Diameter (mm) | Thickness (mm) | Loaded weight (kg) | Gas amount (ml) |
|---|---|---|---|
| 100 | 6 | 1, 2, 3 | 40, 50, 60 |

We mixed oxygen $O_2$ and propane $C_3H_8$ to obtain the combustion gas in this study. $O_2$ and $C_3H_8$ were fully premixed and input into the reaction unit, which enabled the ignitor to activate the combustion. The reaction process leads to a pressure change, which significantly deforms the soft membranes to push the ground. We transformed this physical phenomenon into an equivalent experimental setup, as illustrated in Figure 6(a). The premixed gas was input into the gas chamber and was activated by the ignitor. The resulting combustion deformed the soft membrane and propelled the push-panel upward with a specific load. We used a high-speed camera to record the deformation and recovery of the membrane, and the testing images were converted into binary data. A canny edge algorithm was applied to analyze the experimental results and to calculate the generated $F_t$. In addition, a fatigue experiment was conducted to test the sustainability of the soft membrane, and complete deformation recovery was obtained over hundreds of explosions. The geometric properties and experimental conditions are summarized in Table 2.

We present the axial displacement obtained in the experiment and fitted displacement-time relationships (Figure 6(b)). Having derived the displacement-time functions twice, we obtain the force-time curves, as shown in Figure 6(c). We indicate the nonlinearly fitted density plot of the maximum thrust force with respect to the $O_2$-$C_3H_8$ gas (Figure 6(d)). The entire testing process is illustrated in Movie S5.

The deformation process of a soft membrane caused by the explosion of the $O_2$-$C_3H_8$ gas is illustrated in Figure 6(e). We carried out combustion-thrust experiments to determine the relationship between the gas mix ratio $O_2$-to-$C_3H_8$ $R_{gas}$, the amount of mixed gas $V_{react}$, and the thrust force $F_t$. It can be seen that the membrane expanded from the original shape in "1" to push the panel in "2" during $t = 0.1$-3 ms, which stopped in "3" (explosion limit) at $t = 8.9$ ms. The deformed membrane reduced to "4" at $t = 12$ ms, which eventually left the panel in "5" (contact limit) at $t = 20$ ms. Note that the panel kept moving up because of the thrust force caused by the membrane. Further note that the soft membrane deformed more severely with decreasing $O_2$-to-$C_3H_8$ ratio $R_{gas}$ and increasing $V_{gas}$. The combustion-induced thrust force can be described as as

$$F_t = \Psi \cos \chi t, \quad (3)$$

where $F_t$, $t$, $\Psi$, and $\chi = 180$ rad/s denote the thrust force, time, explosion magnitude, and time period parameter, respectively. See detailed values in Table 3. Experimental results of the thrust testing provide specific data to further numerical simulation and theoretical analysis.

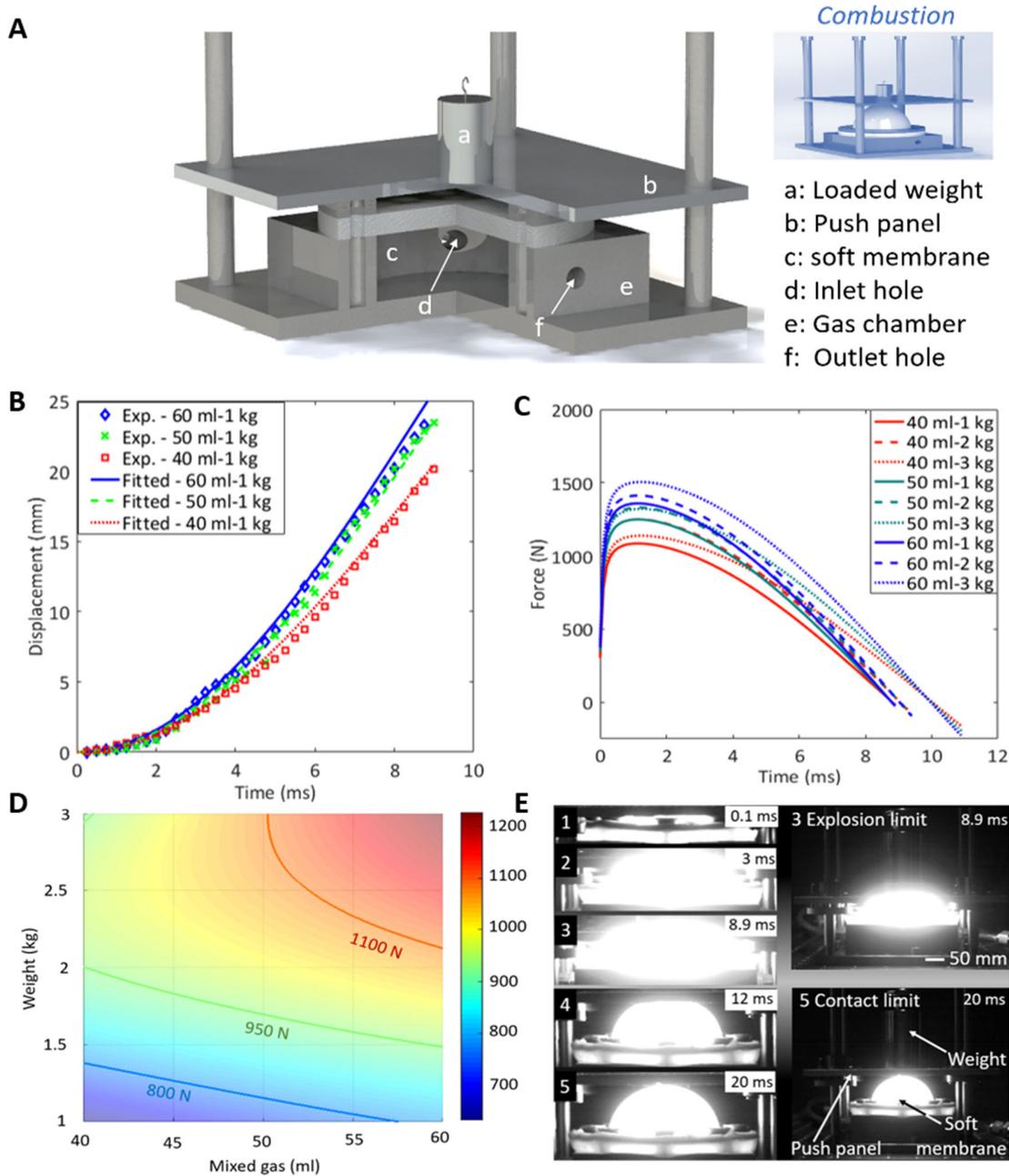

**FIG. 6. (a)** Illustration of the experimental setup and calibration of the soft membrane under the combustion of the $O_2$ and $C_3H_8$ mixed gas. **(b)** Experimental data of the combustion-induced axial displacement and fitted displacement-time curves. **(c)** Thrust force-time relationships for different gas amount-weight ratios. **(d)** Maximum thrust force with respect to the amount of the mixed gas and weight ($O_2$-to-$C_3H_8$ ratio is fixed as 24:5). **(e)** Deformation process of the soft membrane caused by the combustion of the $O_2$-$C_3H_8$ mixed gas. The effective actuation time consumption was about 20 ms, which was divided into two processes: reaching the explosion limit and reaching the contact limit. The expanding membrane pushed the panel upward during the combustion inspiration (i.e., 0.1 ms) and explosion limit (i.e., 8.9 ms), the panel started to decelerate due to the re-dominating of the gravity during the explosion limit and contact limit (i.e., 20 ms).

*Target practice.*

To test the copebot's maneuverability, we carried out target practice experiment under challenging conditions. The target practices are conducted in the reported laboratory-grade testing flume. As illustrated in Figure 7 and shown in Movie

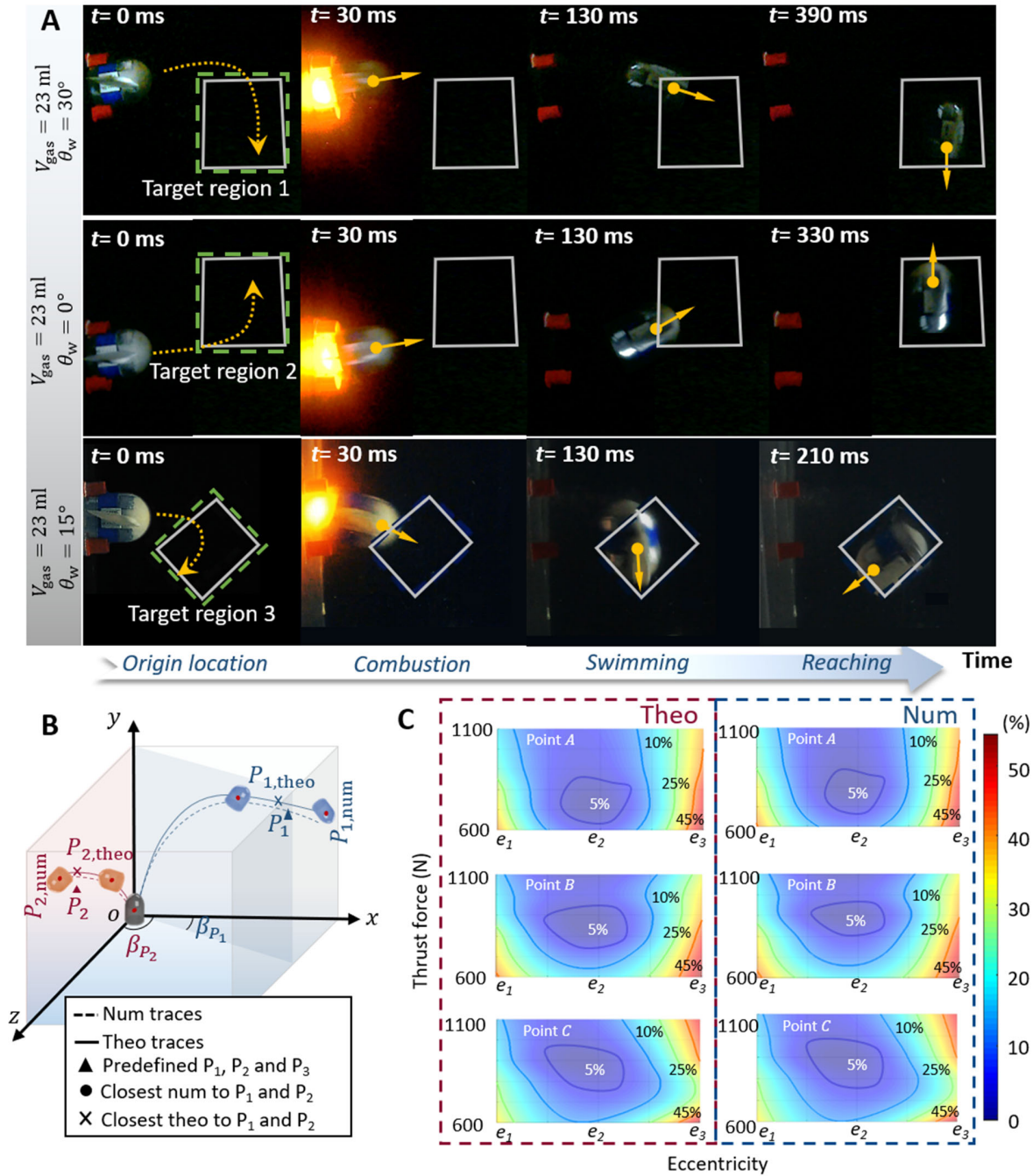

**FIG. 7.** Target practice. **(a)** A copebot prototype jumping horizontally sideways with the purpose to reach predefined targets corresponding to three different motion strategies: an upward quarter turn, downward quarter turn, and 120° downward steering. **(b)** Illustration of the numerical and theoretical target practice. **(c)** Density plots of the numerical and theoretical results.

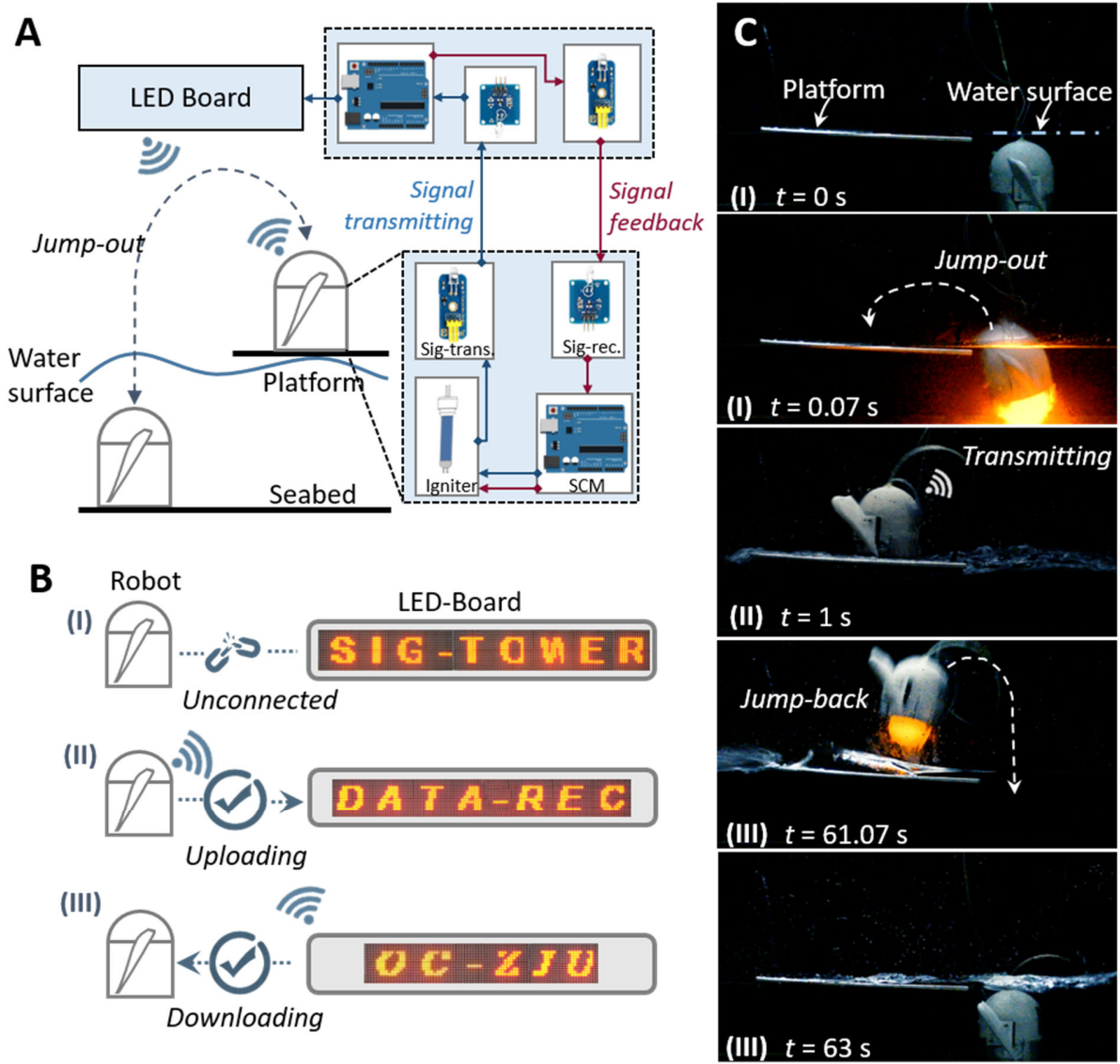

**FIG. 8** Copebot as a signal-transmitting robot. **(a)** Signal transmission strategy. **(b)** Information wirelessly received and transmitted by the signal tower and displayed on a LED screen. "SIG-TOWER", "DATA-REC", and "OC-ZJU" were observed on the screen when the copebot prototype was in following states: not connected, uploaded, and downloaded information, respectively (Movie S4). **(c)** Experimental images of copebot prototype employing the signal transmission strategy.

S3, rather than letting the copebot prototype jump with its torso directed vertically upwards, we let it jump from a wall (where it was initially fixed by the red holders in Figure 7 (a)) with its torso directed horizontally sideways, such that gravity creates an asymmetry. The targets shown as the square frames were predefined on the front walls of the water tank. There are three types of targets showing different motion properties: the upward quarter turn, the downward quarter turn, and the 120° downward turn. All three target types, the copebot prototype was able to reach with a high accuracy. Theoretical and numerical target practices are carried out shown in Figures 7 (b) and (c), and satisfactory agreements are obtained. (See details in Supporting Information).

*Copebot as a signal-transmitting robot. Signal transmission strategy.*

Based on the copebot design, we have built a prototype that is able to wirelessly transmit data through multiphase environments (e.g., water-to-air). As illustrated in Figure 8(a), the robot jumps out of the water and lands on its feet on a small platform above the water. Then, the SCM posts the data to the signal transmitters, which transmit them to a signal tower, which may send data back that can be downloaded into the control unit. When the mission is completed, the robot departs the platform and jumps back into the water for other tasks.

TABLE 3. THRUST FORCES USED IN THE NUMERICAL MODELS AND THE CORRELATED COMBUSTION CONDITIONS.

|  |  | Gas amount (ml) | | |
| --- | --- | --- | --- | --- |
|  |  | 40 | 50 | 60 |
| Weight (kg) | 1 | 630 | 930 | 1050 |
|  | 2 | 730 | 990 | -- |
|  | 3 | 800 | -- | -- |

*Wireless transmission in experiments.*

To build the experimental platform, we designed and fabricated another testing flume with the length-width-height of $2 \times 2 \times 1.2$ m, providing more horizontal space for the leaping motion. Figure 8(c) shows experimental images of a robot using the above signal transmission strategy (see also Movie S4). The LED screen displayed "SIG-TOWER" (I in Figure 8(b)) during $t = 0 - 0.07$ s (step I), which referred to the situation that the robot was not connected to the signal tower. When the robot landed on its feet on the platform during the time period $t = 1 - 61$ s (step II), the LED screen displayed "DATA- REC" (II in Figure 8(b)), which referred to the situation that the robot is successfully connected to the signal tower and that data were received by the tower. After downloading the new information "OC-ZJU" (III in Figure 8(b)) from the signal tower, the robot jumped back into the water (step III).

**Conclusions**

We have designed and built prototypes of the "copebot", an underwater soft robot with a locomotion that mimics that of copepods, which are aquatic animals found in most freshwater and saltwater habitats. The copebot, like copepods, is able to perform explosive curved jumps under water, driven by the combustion of a mixed $O_2$-$C_3H_8$ gas and controlled by adjustable wings attached to the sides of its torso. Parametric studies are conducted to investigate input parameters' sensitivities to kinematic performances of the copebot. We have carried out experiments demonstrating the maneuverability of copebot prototypes under challenging conditions and the potential application of the copebot design for building robots that are able to move forth to land and back to water continuously, while transmitting and receiving data.

The copebot represents a substantial technical advance, since it is able to move up to 8 times faster than previous combustion-driven underwater soft robots, while simultaneously being much more maneuverable. That is, our design addresses the inadequate performance in terms of speed and maneuverability of existing underwater robots and therefore opens the way toward new multifunctional application scenarios such as wireless data transmission in multiphase environments. Apart from mimicking the movability and maneuverability through the combustion-driven method, the reported copebot can be maneuvered by structural optimization (e.g., metamaterials), functional materials (e.g., dielectric materials), and hybrid actuations (e.g., combining combustion with other driving methods).

**Supporting Information**

The data supporting this study are available in the paper and the Supplementary Information. All other relevant source data are available from the corresponding authors upon reasonable request.

Text 1. Numerical simulations of the copebot's motions.
Text 2. Theoretical analysis of the copebot's motions.
Table S1. Geometric and material properties of the numerical models.
Fig. S2. Theoretical analysis of the copebot's motions.
Fig. S3. Relationship of drag coefficient, eccentricity and Reynolds number.
Fig. S4. Repeated traces of the copebot in controllability experiments.
Fig. S5. Trace distributions of the copebot.
Movie S1. The illustration of the automatic pre-inclined wing degree adjustment.
Movie S2. Steering experiments of the copebot captured by high-speed optical capture system.
Movie S3. The copebot reaching predefined targets by different wing angles.
Movie S4. Signal transmitting by continuous jumps.
Movie S5. Thrust testing.

**Author Contributions**

Z.H. conceived the concept and initiated this work. Z.H. and P.J. supervised the project. Y.Y., H.W., G.L. completed the experiments. P.J. and Y.Y. conducted the analytical and numerical models. P.J., Y.Y. and H.W. analyzed the experimental data. Y.Y., P.J. and T.P. wrote the manuscript with contributions from all authors.

**Acknowledgement**

This work was supported by the National Science Foundation of China (Grant no. 11672267), Key Research and Development Plan of Zhejiang, China (2021C03181), Natural Science Foundation of Zhejiang Province (Grant no. LR16E090001) and Fundamental Research Funds for the Central Universities (Grant no. 2017XZZX001-02A). P.J. acknowledges the Startup Fund of the Hundred Talent Program at Zhejiang University.

**Notes**

The authors declare no competing financial interest.